# Exploiting Transductive Property of Graph Convolutional Neural Networks with Less Labeling Effort

1st Yasir KILIÇ
*Computer Engineering Department*
Adana Alparslan Turkes Science
and Technology University
Adana,Turkey
ykilic@atu.edu.tr

ORCID: 0000-0001-9666-3746

*Abstract—*

**Recently, machine learning approaches on Graph data have become very popular. It was observed that significant results were obtained by including implicit or explicit logical connections between data samples that make up the data to the model. In this context, the developing GCN model has made significant experimental contributions with Convolution filters applied to graph data. This model follows Transductive and Semi-Supervised Learning approach. Due to its transductive property, all of the data samples ,which is partially labeled, are given as input to the model. Labeling, which is a cost, is very important. Within the scope of this study, the following research question is tried to be answered: If at least how many samples are labeled, the optimum model success is achieved? In addition, some experimental contributions have been made on the accuracy of the model, whichever sampling approach is used with fixed labeling effort. According to the experiments, the success of the model can be increased by using the local centrality metric.**

*Keywords—graph convolution, transductive learning, graph, GCN (graph convolutional neural networks) ,*

## I. INTRODUCTION

The data is by nature built by implicit or explicit connected entities and their features. The data is widely modeled as tabular. However, since the modeling is weak the relationship between the entities in the data, graph-based modelling and related learning algorithms are studied in recently.

Attractive results were obtained by applying the Convolution process, which has recently been applied to image data, to graph-based data. In this context, the proposed GCN model is the basis of such studies. In this study, experiments are conducted on how model success can be increased by using semisupervised and transductive features of this model. Also, the developing GCN model has made significant experimental contributions with Convolution filters applied to graph data.

The fact that the model is semi-supervised [1] [2] is that some of the samples are labeled and transductive [3], while the unlabeled data is included in the model at the time of training. In this study, experiments have been carried out as to whether these features of the model can be exploited. Although the results obtained are not very strong, they contain experimental indications that the approach can be successful.

## II. RELATED WORKS

This section organizes literature review about graph data-based models and their learning approaches.

Graph based learning models are generally researched in two categories as explicit Graph Laplacian Regularization and Graph Embedding approaches. In Graph Laplacian Regularization, although there are a lot of studies; label propagation [4], manifold regularization [5] and semi-supervised embedding but not graph [6]. Since these approaches are based on looking at the neighborhood properties of the samples, they are inadequate in modeling the data according to the graph-based embedding approaches [2]. Graph Embedding studies, on the other hand, aim to produce a representation vector that expresses the whole graph structure with neighborhood information. To implement this studies, pretraining should be done. This is a cost for construction of model. For less cost and simplicity, we exclude graph embedding pretraining approach.

In literature, semi-supervised and transductive features of the GCN model have been researched from various view. [7] revisits semisupervised property of GCN and then proposes inductive and transductive formulation for the model. [1] trying to find best serve graph architectre for semi-supervised GCN model. These studies did not examine the effect of initial labeled nodes on the accuracy of the model.However , this study observes effect of initial labelled samples for GCN model. We follow experimental settings in [2] with raw feature representation and given hyperparameters. To the best of our knowledge, our study is first research to exploit transductive property of GCN and related analysis.

### III. MATERIAL AND METHODS

In this section, definitions and descriptions about materials(e.g dataset, programming language,software library) and methods used in this study .

#### A. Materials

Within the scope of this study, 3 different citation network datasets are used, which are widely used in the literature: CORA, Citeseer, Pubmed [8] . The statistical properties of these datasets are listed in Table 1.

*Table 1. Statistical Properties of Used Datasets*

| Dataset | #nodes | #edges | #CC | #classes | #features |
|---|---|---|---|---|---|
| CORA | 2708 | 5429 | 78 | 7 | 1433 |
| CITESEER | 3312 | 4732 | 438 | 6 | 3703 |
| PUBMED | 19717 | 44338 | 1 | 3 | 500 |

The experiments implemented in this study are run using open source Python programming language. Graph-based data were analyzed using the NetworkX library [9]. Also, StellarGraph is a library which are supports graph based machine learning models such as GCN. To implement GCN, this library was used. In addition, Google Colab [10] coding environment was used due to efficient memory and processing capacity.

#### B. Graph Convolutional Neural Networks(GCNs)

GCNs are a proposed deep machine learning approach to learn graph based data using Convolution operation [2]. The most important part is the propogation rule that graph data will follow during the learning phase. This rule determines the function of propagating information to other nodes in each layer of the model and is formulated as follows:

$$H^{(l+1)} = \sigma\left(D^{-0.5} A\, D^{-0.5} H^{(l)} W^{(l)}\right) \quad (1)$$

Where σ denotes non-linear activation functions such as ReLU, D and A are matrices that are diagonal degree matrix and self-adding adjacency matrix respectively. [2] shows that this propagation rule is motivated by first order approximation of localized spectral filters on graphs [11] [12].

## C. GCNs As A Tranductive Semi-Supervised Learning

Let we have labelled and unlabeled samples. Semi-Supervised learning generate a learning model using these labelled and unlabelled samples, In this point, there are two commonly used learning paradigms: Inductive and Transductive Learning.

Inductive learning is aims to generalize model for unseen samples , whereas, Transductive learning is only aims to classify unlabeled samples which seen during training phase. In transductive learning, the model does not generalize unseen samples .

The GCN model follows semi-supervised learning transductive paradigm. Since this paradigm is used, during the training phase, the loss function is calculated both according to the tagged samples and the unlabeled connected samples [2].

## D. Spectral Graph Clustering

The eigen values of Graph Laplacian Matrix indicate important properties about graphs. One of the most important of these features is the connectivity of graph nodes [13] [14]. Graph Laplacian Matrix is a matrix that represents regularization across nodes [4]. Graph Laplacian Matrix formula as follows:

$$L = D^{-0.5}(D - A)D^{-0.5} \quad (2)$$

Where D denotes diagonal degree matrix of any graph and A represents weighted or not adjacency matrix. Then , Eigenvalues and eigenvectors of the matrix gives algebraic co-boundary between nodes (See Eq 3). Using this eigen values, spectral clustering on graph algorithm is performed [14] [13].

$$L\,x = \lambda x \quad (3)$$

## E. Centrality : Node Importance Metric

According to Graph Theory, centrality is a metric that measure node importance across whole graph. Although this metric determines relatively importance on the global graph, it is meaningless especially in disconnected graphs.

In this study, a local centrality metric [15] ,which is local reaching centrality, is used because there may be a non-connected graph and maximum $2^{nd}$ order convolution filter is implemented. The formula of the local reaching centrality metric used is as follows:

$$C_R(i) = \frac{1}{N-1} \sum_{j:0<d(i,j)<\infty} \frac{1}{d(i,j)} \quad (4)$$

Where N denotes total number of nodes in connected graph space and d(i,j) represents distance between node i and j.

## IV. EXPERIMENTS

This section makes quantitavely analysis over transductive property of Graph Convolutional Neural Networks. Basically, a controlled experiment is carried out and it is analyzed whether it is possible to make the transductive feature of the GCN model effective using the graph centrality.

In the first experiment, the distribution is extracted to get information about the distribution of the positions of the samples in the data sets used on the diagram. In this experiment, analysis of the spectrum of the graph laplacian matrix is performed. These values show the connectivity properties of the graph nodes. Negative and smaller values of eigen values in CORA and CITESEER data indicates lower graph connectivity. But PUBMED data is more connected.

*Table 2 . Statistical Properties of Graph Normalized Laplacian Eigenvalues*

|        | CORA            | CITESEER        | PUBMED       |
|--------|-----------------|-----------------|--------------|
| min    | -3.9178412e-06  | -5.1841025e-06  | 6.198883e-06 |
| median | 1               | 0.999           | 0.999        |
| avg    | 1               | 0.978           | 0.999        |
| std    | 0.526           | 0.670           | 0.367        |
| max    | 2               | 2               | 1.985        |

In the second experiment, the default labeling rate is fixed, and the results are listed under 4 different sample selection policies (DF: Default Sample Selection, MC: Most Central Sample Selection, LC: Least Central Sample Selection and ECM: Equal Combined Central Sample Selection). Three different metrics are used to evaluate policies: Acc (Test Accuracy), Loss (Categorical Cross Entropy Loss), Stop (Epoch Number at Early Stopping). The proposed approach, the ECM model, uses half the labeling capacity to the node with a high local centrality and the other half to the lowest. It is observed that this approach is more successful than other models. This model uses half of the cost to label the most effective(central) nodes to cover the graph and the other half to the nodes where the flow of information is limited (less central). Other models appear to be overfit in a shorter time (see Stop metric), meaning the data is memorized and the same classification estimates are always made.

*Table 3. Model Evaluation Under Fixed Default Labelling Rate*

|          | **Labeling Policy** | *MC*  |       |      | *LC*  |           |      | *ECM*     |       |        |
|----------|---------------------|-------|-------|------|-------|-----------|------|-----------|-------|--------|
| **Dataset** | **Labeling Rate** | Acc   | Loss  | Stop | Acc   | Loss      | Stop | Acc       | Loss  | Stop   |
| *CORA*     | *0.053*           | *0.769* | *0.83* | *29* | *0.775* | ***0.759*** | *39* | ***0.777*** | *0.792* | ***51*** |
| *CITESEER* | *0.036*           | *0.574* | *1.364* | *21* | *0.564* | ***1.327*** | *23* | ***0.593*** | *1.453* | ***37*** |
| *PUBMED*   | *0.003*           | *0.781* | *0.658* | *48* | *0.7*   | *1.075*   | *15* | ***0.787*** | *0.67*  | ***55*** |

In the third and final experiment, the performances of the proposed methods according to the changing labeling cost were measured using the Accuracy metric. In fact, it cannot be said that a model is better when looking at the experiments. In fact, when the labeling rate of the models increases, success is expected to increase linearly. When the point where the success starts to decrease is considered as the saturation point, it is seen that the models reach saturation at approximately the same points (underlined results). However, the stability of the models shows important clues about the robustness of the models. Looking at the disconnected graph data (Figures 4.1 and 4.2), it is seen that the proposed ECM approach gives more stable results.

*Table 4. Model Evaluation with Respect To Labelling Rate*

| | **CORA** | | | | **CITESEER** | | | | **PUBMED** | | |
|---|---|---|---|---|---|---|---|---|---|---|---|
| Labeling Rate | | MC | LC | ECM | | MC | LC | ECM | | MC | LC | ECM |
| *0.05* | | **0.768** | *0.759* | *0.759* | | *0.609* | **0.613** | *0.6* | | *0.707* | *0.844* | *0.844* |
| *0.1* | | *0.778* | *0.784* | **0.8** | | *0.616* | **0.633** | *0.62* | | **0.856** | **0.856** | *0.854* |
| *0.15* | | *0.787* | **0.805** | *0.8* | | *0.647* | <u>0.625</u> | *0.637* | | <u>**0.854**</u> | <u>0.738</u> | <u>0.661</u> |
| *0.2* | | **0.814** | *0.806* | *0.8* | | <u>0.625</u> | **0.661** | *0.657* | | *0.858* | *0.858* | **0.861** |
| *0.25* | | <u>0.805</u> | *0.811* | **0.817** | | *0.651* | **0.655** | <u>0.636</u> | | *0.861* | **0.864** | *0.661* |
| *0.3* | | **0.813** | <u>0.808</u> | *0.8* | | *0.659* | *0.674* | *0.665* | | **0.862** | *0.629* | *0.852* |
| *0.35* | | *0.817* | *0.816* | *0.816* | | *0.663* | *0.65* | **0.665** | | *0.866* | *0.652* | *0.866* |
| *0.4* | | *0.821* | *0.809* | *0.82* | | *0.637* | **0.69** | *0.675* | | *0.865* | *0.72* | **0.868** |

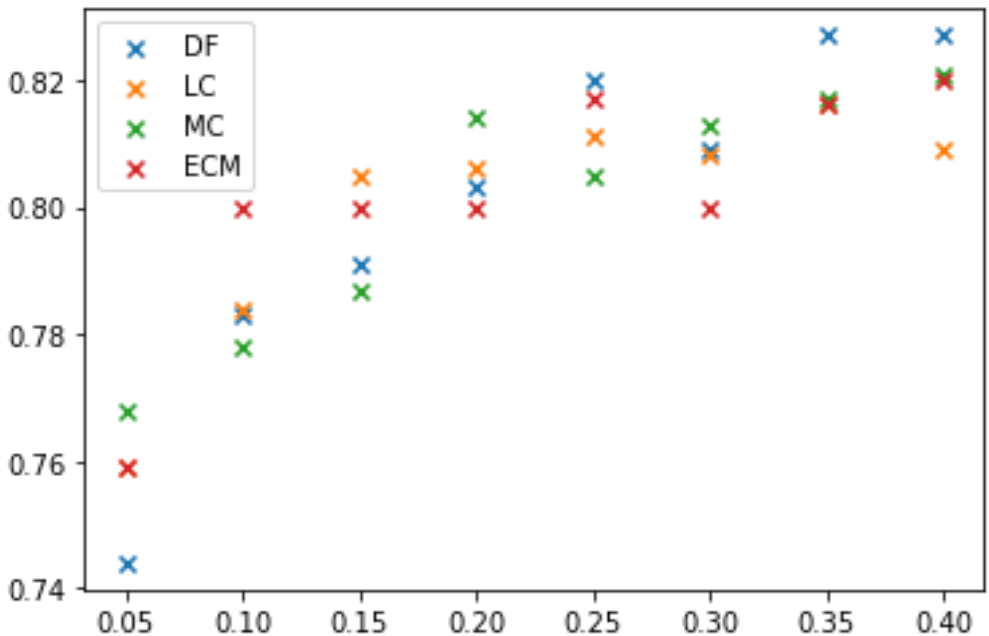

*Figure IV.1 Changing Labelling Rate Performance on CORA DATASET*

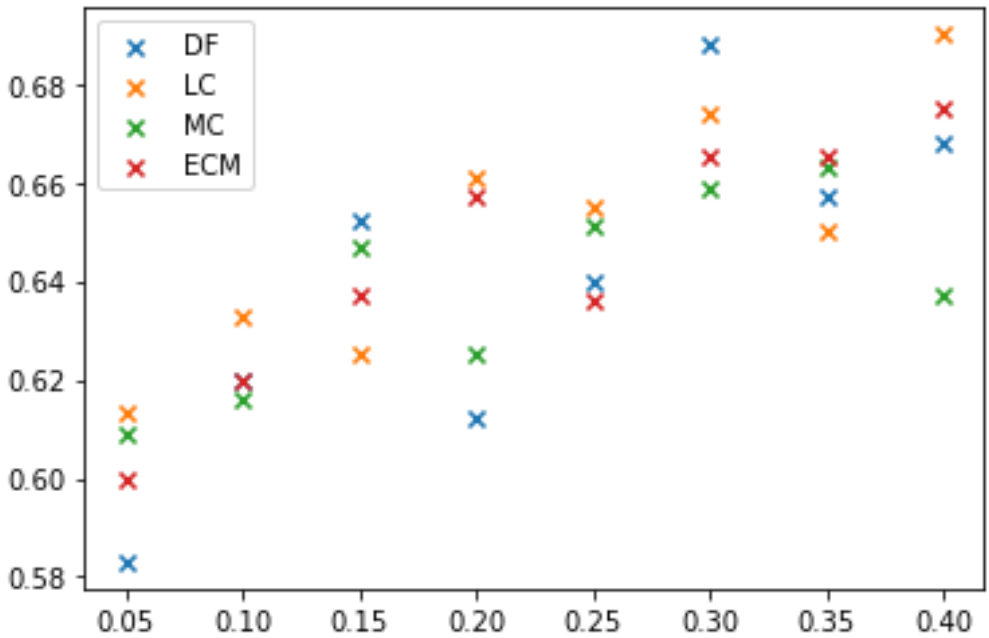

*Figure IV.2 Changing Labelling Rate Performance on CITESEER DATASET*

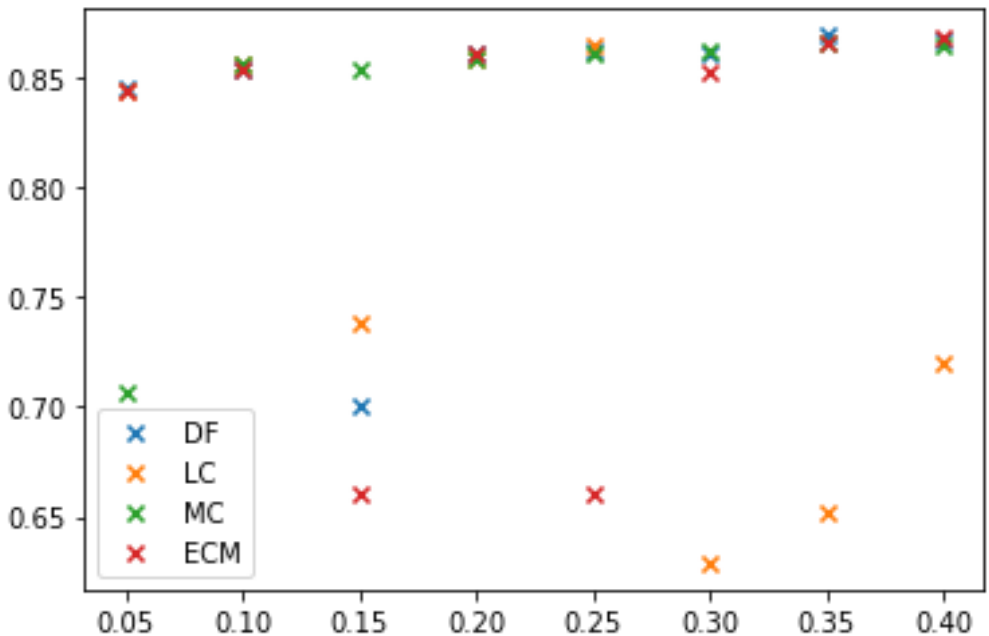

*Figure IV.3 Changing Labelling Rate Performance on PUBMED DATASET*

If we summarize the experiments, generally 2 controlled experiments are done. These experiments observe the success of the model by fixing the labeling cost and the success of the model is observed according to the change of the labeling cost. It is observed that the proposed ECM model achieves successful results especially on disconnected graphs. In the future, it is considered that this study will be developed with a more comprehensive formulated labeling approach.

## V. CONCLUSIONS

An approach to improve their performance using the semisupervised and transductive features of the GCN model, which is the basis of the studies on the application of the Convolution process on Graph data, is proposed. Since the samples are labelled during the training and which data are very important, this study shows experimentally how to increase the success of the GCN model using the local centrality approach. In the future, this work can be extended from several perspectives:
- Choosing training data according to the graph laplacian spectrum
- A new approach should be proposed for disconnected graph learning problem.

## VI. REFERENCES

[1] B. Jiang,, . Z. Zhang, D. Lin, J. Tang and B. Luo, "Semi-supervised Learning with Graph Learning-Convolutional Networks," in *Proceedings of the IEEE Conference on Computer Vision and Pattern Recognition*, 2019.

[2] T. N. Kipf and M. Welling, "SEMI-SUPERVISED CLASSIFICATION WITH GRAPH CONVOLUTIONAL NETWORKS," *arXiv preprint arXiv:1609.02907,* 2016..

[3] Wikipedia, "Transduction_(machine_learning)," 2020. [Online]. Available: https://en.wikipedia.org/wiki/Transduction_(machine_learning).

[4] X. Zhu, Z. Ghahramani and J. Lafferty, "Semi-Supervised Learning Using Gaussian Fields and Harmonic Functions," in *International Conference on Machine Learning (ICML)*, 2003.

[5] "Manifold Regularization:A Geometric Framework for Learning from Labeled and Unlabeled Example," *Journal of Machine Learning Research,* no. 7, 2006.

[6] J. Weston, F.́́. Ratle and R. Collobert, "Deep Learning via Semi-Supervised Embedding," in *Neural networks: Tricks of the trade*, Berlin, Springer, 2012, pp. 639-655.

[7] Z. Yang, W. W. Cohen and R. Salakhutdinov, "Revisiting semi-supervised learning with graph embeddings," *arXiv preprint,* no. arXiv:1603.08861., 2016.

[8] P. Sen , G. Namata, M. Bilgic , L. Getoor , B. Galligher and T. Eliassi-Rad , "Collective Classification in Network Data," *AI Magazine,* vol. 29, no. 3, 2008.

[9] NetworkX developers, "NetworkX: Network Analysis in Python," NetworkX developers, 2014. [Online]. Available: https://networkx.github.io/. [Accessed 2020].

[10] GOOGLE INC, "Google Colaboratory," Google, [Online]. Available: https://colab.research.google.com. [Accessed 2020].

[11] D. K. Hammond, P. Vandergheynst and R. Gribonval, "Wavelets on graphs via spectral graph theory," *Applied and Computational Harmonic Analysis,* vol. 30, no. 2, pp. 129-150, 2011.

[12] M. Defferrard, X. Bresson and P. Vandergheynst, "Convolutional neural networks on graphs with fast localized spectral filtering," *Advances in neural information processing systems,* pp. 3844-3852, 2016.

[13] J. Shi and J. Malik, "Normalized Cuts and Image Segmentation," *IEEE TRANSACTIONS ON PATTERN ANALYSIS AND MACHINE INTELLIGENCE,* vol. 22, no. 8, 2000.

[14] S. X.Yu and J. Shi, "Multiclass Spectral Clustering," *ieeexplore.ieee.org,* 2003.

[15] E. Mones, L. Vicsek and T.́. Vicsek, "Hierarchy Measure for Complex Networks," *PloS one,* no. 7.3, 2012.

[16] Data61, CSIRO Revision 61b17d9f, "Welcome to StellarGraph's documentation!," Data61, CSIRO Revision 61b17d9f, [Online]. Available: https://stellargraph.readthedocs.io/en/latest/demos/interpretability/index.html. [Accessed 2020].

[17] D. Zhou, . O. Bousquet, T. N. Lal, J. Weston and B. Schölkopf , "Learning with local and global consistency," in *Advances in Neural Information Processing Systems 16*, 2004.

[18] T. Joachims, "Transductive Inference for Text Classification using Support Vector Machines," *Icml,* no. 99, pp. 200-209, 1999.